%% file: main.tex
\documentclass[10pt,twocolumn,letterpaper]{article}

\usepackage[algorithms]{wacv}      

\usepackage{graphicx}
\usepackage{amsmath}
\usepackage{amssymb}
\usepackage{booktabs}
\usepackage{algorithm}
\usepackage{algorithmic}
\input{math}

\usepackage{multirow, float, booktabs, boldline, hhline}
\usepackage{enumitem}
\usepackage{pifont}

\AtBeginDocument{%
  \providecommand\BibTeX{{%
    \normalfont B\kern-0.5em{\scshape i\kern-0.25em b}\kern-0.8em\TeX}}}

\usepackage[pagebackref,breaklinks,colorlinks]{hyperref}

\usepackage[capitalize]{cleveref}
\crefname{section}{Sec.}{Secs.}
\Crefname{section}{Section}{Sections}
\Crefname{table}{Table}{Tables}
\crefname{table}{Tab.}{Tabs.}

\begin{document}

\title{FastEdit: Fast Text-Guided Single-Image Editing via \\ Semantic-Aware Diffusion Fine-Tuning}

\author{Zhi Chen\footnotemark[1]\footnote{aaa}, Zecheng Zhao\footnotemark[1], Yadan Luo, Zi Huang\\
University of Queensland\\
\tt\small*Equal Contribution\\
\tt\small\{zhi.chen, jason.zhao, y.luo, helen.huang\}@uq.edu.au
}


\twocolumn[{%
\renewcommand\twocolumn[1][]{#1}%
\maketitle

\renewcommand{\thefootnote}{\fnsymbol{footnote}} 
\footnotetext[0]{Equal contribution.} 

\includegraphics[width=\linewidth]{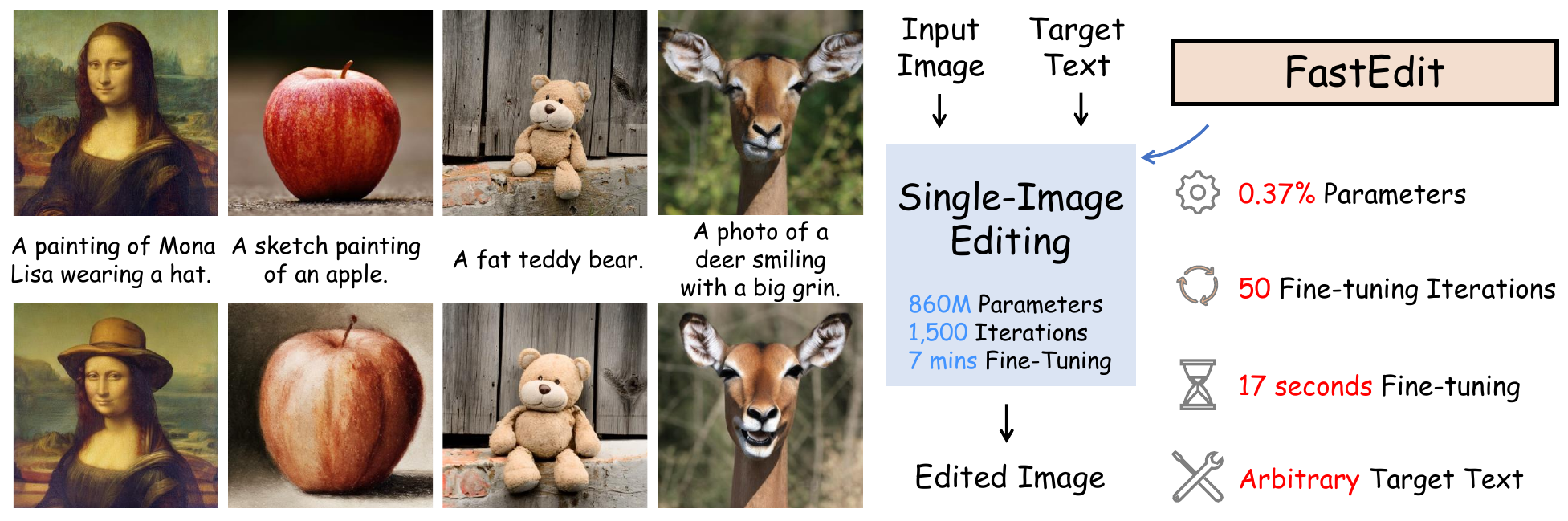}
\captionof{figure}{FastEdit – Text-Guided Single-Image Editing in 17 seconds. We show the pairs of 512$\times$512 input images, and the given target texts with corresponding edited results. Compared with the baseline methods, FastEdit fine-tunes only 0.37\% parameters for 50 iterations. Arbitrary target texts are supported for the fine-tuned model, due to its embedding optimization-free nature.}
\label{fig:teaser}
\vspace{10pt}
}]


\begin{abstract}
Conventional Text-guided single-image editing approaches require a two-step process, including fine-tuning the target text embedding for over 1K iterations and the generative model for another 1.5K iterations. 
Although it ensures that the resulting image closely aligns with both the input image and the target text, this process often requires 7 minutes per image, posing a challenge for practical application due to its time-intensive nature. 
To address this bottleneck, we introduce FastEdit, a fast text-guided single-image editing method with semantic-aware diffusion fine-tuning, dramatically accelerating the editing process to only 17 seconds. 
FastEdit streamlines the generative model's fine-tuning phase, reducing it from 1.5K to a mere 50 iterations. 
For diffusion fine-tuning, we adopt certain time step values based on the semantic discrepancy between the input image and target text. 
Furthermore, FastEdit circumvents the initial fine-tuning step by utilizing an image-to-image model that conditions on the feature space, rather than the text embedding space. 
It can effectively align the target text prompt and input image within the same feature space and save substantial processing time. 
Additionally, we apply the parameter-efficient fine-tuning technique LoRA to U-net. With LoRA, FastEdit minimizes the model's trainable parameters to only 0.37\% of the original size. At the same time, we can achieve comparable editing outcomes with significantly reduced computational overhead. 
We conduct extensive experiments to validate the editing performance of our approach and show promising editing capabilities, including content addition, style transfer, background replacement, and posture manipulation, \textit{etc}. Project page: \url{https://fastedit-sd.github.io/}
\end{abstract}

\section{Introduction}
\begin{figure*}[h!]
    \centering
    \includegraphics[width=0.97\linewidth]{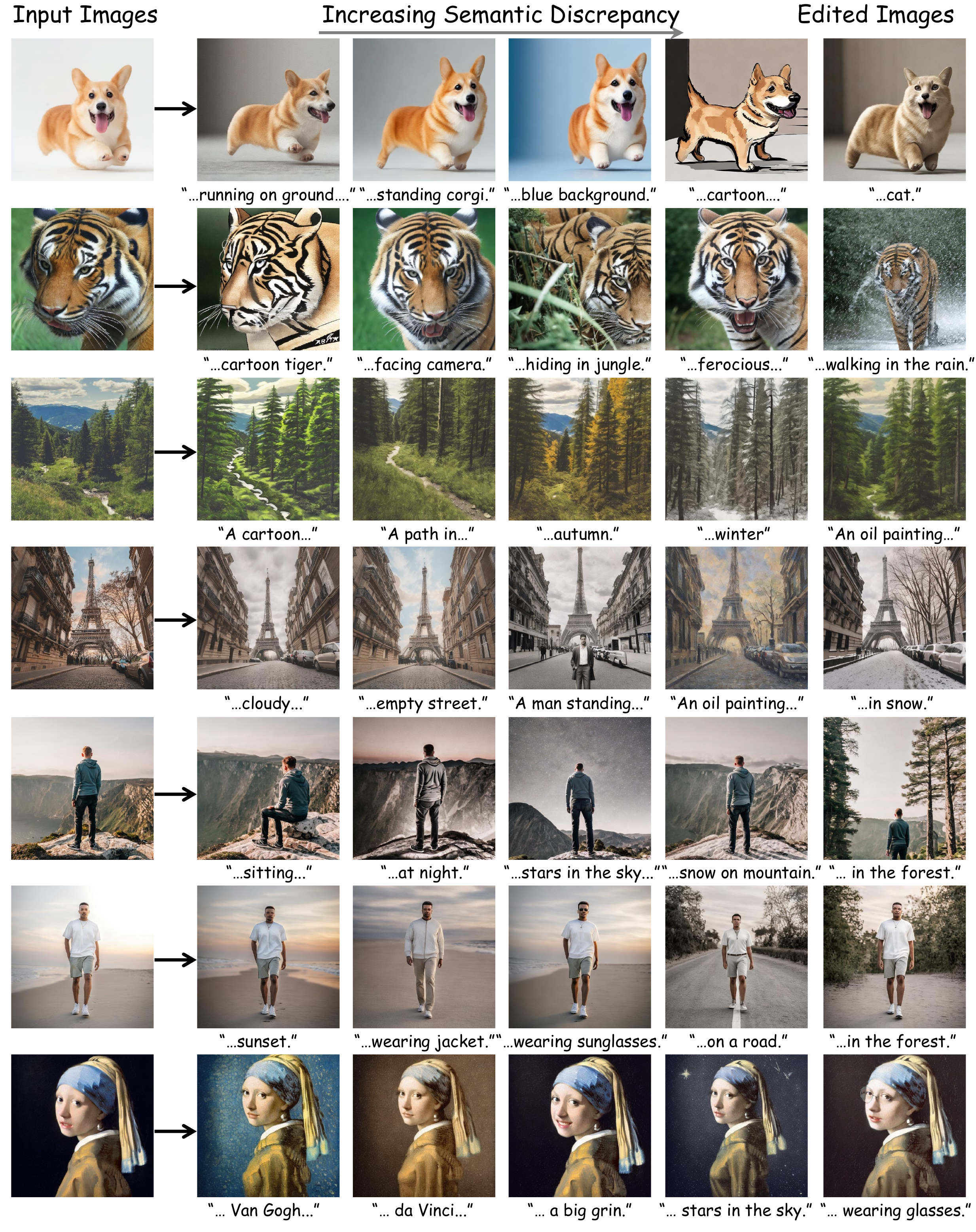}
    \vspace{-1.5em}
    \caption{Different target texts applied to the same images. FastEdit edits the same image differently based on the semantic discrepancy between input image and the target texts.}
    \label{fig:multi_texts}
    \vspace{-10pt}
\end{figure*}

The exponential rise in digital content creation \cite{arnaud2006collada,earnshaw2012digital} has significantly amplified the demand for advanced image editing software, such as Adobe Photoshop and Canva Photo Editor. The editing of visual content traditionally demands substantial manual input, being both labor-intensive and requiring a high degree of skill and artistic judgment, often necessitating prolonged periods of experimentation to achieve the desired outcome \cite{saad2008easy,knochel2016photoshop}. 
Deep learning has revolutionized the field of image editing by introducing various models capable of understanding and manipulating visual content in ways that mimic and sometimes outperform human capabilities \cite{portenier2018faceshop,patashnik2021styleclip}. Unlike traditional image editing tools that rely on manual adjustments and a deep understanding of photo editing software, deep learning models can automatically perform complex tasks such as enhancing image resolution, correcting lighting and color imbalances, or even generating realistic content within a photo. 

Recently, single-image editing has attracted much attention with the photorealistic generation power in generative models \cite{zhu2020domain,cheng2020sequential,abdal2021styleflow,abdal2022clip2stylegan}. It involves modifying a specific image without using multiple images to provide a context \cite{oh2001image}. Particularly, recent large-scale text-to-image diffusion models, \textit{e.g.,} DALL·E \cite{ramesh2022hierarchical}, Stable Diffusion \cite{rombach2022high}, and Imagen \cite{saharia2022photorealistic}, have greatly enhanced the image editing performance. This is particularly useful when the image editing could be guided by a simply natural language text prompt. There are many approaches developed for text-guided single-image editing \cite{gal2022stylegan,abdal2022clip2stylegan,kawar2023imagic,han2023svdiff}, demonstrating impressive editing quality.


Imagic \cite{kawar2023imagic} is the current state-of-the-art approach to text-guided single-image editing. This method optimizes the target text embedding while keeping the diffusion model architecture frozen. Through this process, it is possible to derive an input text embedding that is closely aligned with the desired modifications. Subsequently, the denoising model is fine-tuned using both the newly obtained text embedding and the original image. This strategy ensures the preservation of the overall appearance of the image while introducing the specified alternation. As a result, when provided with a target text embedding, the fine-tuned model is capable of generating a new image that embodies minor adjustments requested in the text, yet maintains the visual continuity of the original image. 

Despite its impressive image editing performance, the editing time costs pose a significant challenge. To edit a single image, it typically costs approximately 7 minutes. To accelerate the editing process, in this paper, we propose a fast text-guided single-image editing method, \textbf{FastEdit}, that reduces a single-image editing time to $\sim$17s while keeping the editing quality.

First, we introduce semantic-aware diffusion fine-tuning, which significantly reduces the number of training iterations required to just 50. Specifically, in the inversion process of diffusion models, a time step determines the noise added to the input image. The time step is uniformly sampled between 1 and 1000, which is a large range. Thus, fine-tuning a diffusion model typically requires a large number of training iterations (1,500 for Imagic) to accommodate various time steps. This is the core challenge that hinders the acceleration of the editing process. Our method leverages observations indicating that smaller time step values result in the model learning minor image modifications, while larger values induce more substantial changes in image structure or composition. By assessing the semantic discrepancy between the input image and target text, we adeptly select time step values that align with the necessary degree of change.
This approach, illustrated in our semantic discrepancy analysis in {\color{red}{Fig. \ref{fig:semantic_discrepancy}}}, streamlines the fine-tuning process to require dramatically fewer iterations.
As a result, our semantic-aware strategy enables efficient single-image editing within just 50 iterations, offering a 30-fold increase in efficiency compared to traditional methods. 
This advancement represents a significant leap forward in optimizing the text-guided image editing process, making it faster and more responsive to the desired semantic edits.


Second, we effectively bypass the initial embedding optimization step, usually consuming around 2 minutes of fine-tuning time. Our strategy employs the image-to-image variant of the Stable Diffusion model rather than the text-to-image variant used in Imagic. The text-to-image approach has a notable limitation: it cannot fully leverage CLIP's ability to align textual and visual features.
Specifically, it conditions on tokenized text embeddings from CLIP, which may not align with the visual modality in the same way that textual features processed by CLIP's text encoder do. By contrast, the image-to-image variant is conditioned directly on CLIP's image features. Thus, by utilizing CLIP's image encoder, we directly extract image features to serve as the condition for generating the input image.

Third, in our quest to further expedite the fine-tuning process, we delve into reducing the model's trainable parameters by incorporating parameter-efficient fine-tuning techniques such as prompt-tuning \cite{zhou2022learning}, adapters \cite{gao2024clip}, and Low-Rank Adaptation (LoRA) \cite{hu2021lora}. While our experimentation reveals that both prompt-tuning and adapters encounter challenges related to language drift \cite{ruiz2023dreambooth}, LoRA stands out by delivering performance comparable to full fine-tuning but with a drastic reduction in trainable parameters (only 0.37\%). This significant reduction in parameters not only streamlines the editing process but also maintains high-quality outcomes, underscoring LoRA's effectiveness in balancing efficiency with performance. 
To summarize, the contribution of this paper is threefold:
\begin{itemize}[align=parleft,left=0em]
\item{We introduce a fast text-guided single-image editing approach, namely FastEdit, which can perform complex non-rigid edits on a single image in $\sim$17s on a single A6000 GPU. }
\item{We propose a semantic-aware diffusion fine-tuning strategy that requires only 50 fine-tuning iterations for a single-image editing. Specifically, we determine the values of the diffusion time step condition based on the semantic discrepancy between the input image and the target text.}
\item{Extensive experiments conducted on our collected image-text pairs and the image editing benchmark TEdBench demonstrates the efficiency and effectiveness of our proposed FastEdit in the text-guided image editing task.}
\end{itemize}

\section{Related Work}
\textbf{Text-Guided Image Editing.}
Text-guided image editing with generative models represents a transformative approach in digital content creation, where textual descriptions, rather than manual interventions, steer the alterations made to an image \cite{abdal2022clip2stylegan, abdal2021styleflow, patashnik2021styleclip, harkonen2020ganspace, li2020manigan}. 
Recent advances in conditional diffusion models \cite{ho2020denoising,sohl2015deep,song2020denoising,rombach2022high,ni2023degeneration,deng2023mv} have significantly transformed the landscape of text-guided image editing, offering new solutions that yield impressive results \cite{kawar2023imagic,gal2022image,ruiz2023dreambooth,nichol2022glide}.
GLIDE \cite{nichol2022glide} is the first text-guided image model that leverage text for directly guiding image synthesis at the pixel level. Similar to GLIDE, Imagen \cite{saharia2022photorealistic} employs a cascaded framework in multiple stages to generate high-resolution images more efficiently in pixel space. Following the GAN-based methods \cite{gal2022stylegan, abdal2022clip2stylegan} that employs CLIP \cite{radford2021learning} for guiding image editing with text, many studies incorporate CLIP into diffusion models. A representative work is DiffusionCLIP \cite{kim2022diffusionclip}. With the power of CLIP, DiffusionCLIP enables image manipulation in both seen and unseen domains. SDEdit \cite{meng2021sdedit} leverages a diffusion model with a generative prior rooted in stochastic differential equations (SDEs) to create realistic images through iterative denoising. 
UniTune \cite{valevski2022unitune} introduces a novel approach to fine-tuning diffusion models by focusing on a single base image during the tuning phase. 
Some approaches concentrate on partially fine-tuning specific components of the denoising model.
An exemplar of this strategy is KV Inversion \cite{huang2023kv}, which innovates by optimizing the learning of keys (K) and values (V) within the model. 
Gal et al. \cite{gal2022image} proposed text inversion that allows for the personalization of text-to-image generation models without requiring direct access or modifications to the model's parameters. It focuses on adapting the way textual prompts are interpreted by the model.
Dreambooth \cite{ruiz2023dreambooth} further introduces a class-specific prior preservation loss that addresses the language drift problem in the text inversion method. 
Imagic \cite{kawar2023imagic} is the current state-of-the-art approach to single-image editing. This method focuses on fine-tuning the target text embedding while keeping the core model architecture frozen. 
Subsequently, the denoising model is fine-tuned using both the new obtained text embedding and the original image. 
As a result, when provided with a target text embedding, the fine-tuned model is capable of generating a new image that embodies minor adjustments requested in the text.
Despite its image editing performance, its editing time cost pose a significant challenge. It requires $\sim$7 minutes to edit a single image. In this work, we reduce the time to around 17 seconds for single-image editing.

\noindent\textbf{Parameter-Efficient Fine-tuning.} 
The emergence of large pre-trained foundation models has revolutionized various fields, including CLIP \cite{radford2021learning}, GPT-3 \cite{brown2020language}, Stable Diffusion \cite{rombach2022high}, and LLAMA \cite{touvron2023llama}. A common approach to leverage these models for specific tasks involves fine-tuning to adapt to the target task. However, this strategy often encounters challenges when the target task is associated with small datasets, leading the fine-tuned model to overfitting to the dataset. To cope with this challenge, many parameter-efficient fine-tuning methods are proposed, including prompt tuning \cite{zhou2022learning,jia2022visual,shu2022test}, adapter tuning \cite{gao2024clip,alayrac2022flamingo,najdenkoska2023meta}, LoRA \cite{hu2021lora,smith2023continual,yu2021differentially}, \textit{etc}.
Prompt tuning models customization by learning a context embedding in the input data. CoOp \cite{zhou2022learning} pioneers this method by learning the embeddings of prefix prompts in CLIP \cite{radford2021learning}. Following a similar vein, VPT  \cite{jia2022visual} unveils a visual prompt tuning method that subtly introduces a small set of parameters directly into the image input space. Additionally, TPT \cite{shu2022test} explores test-time tuning, a technique designed to dynamically learn adaptive prompts with just a single test sample. 
CLIP-Adapter \cite{gao2024clip} incorporates adapters into both the image and text encoders of the CLIP framework to fine-tune the output representations.  Similarly, Flamingo \cite{alayrac2022flamingo} extends the versatility of CLIP by bridging its image encoder with large language models (LLMs). This integration enables Flamingo to efficiently handle complex interactions between text and visual inputs. Moreover, ML-MFSL \cite{najdenkoska2023meta} applies the power of LLMs to the domain of meta-learning. Llama-Adapter\cite{zhang2023llama} introduce adapters to LLAMA which accelerate the large language model fine-tuning.
LoRA \cite{hu2021lora} model adaptation through the introduction of a low-rank matrix designed to adjust the weight matrices of specific layers within a pre-trained model. LoRA lsignificantly reduces the size of learnable parameters compared to the original weight matrices. Building upon this, c-LoRA \cite{smith2023continual} takes the concept further by introducing a continually self-regulated low-rank adaptation specifically for the cross-attention layers of the widely used Stable Diffusion model. 
DP-LoRA \cite{yu2021differentially} addresses the critical aspect of privacy in the fine-tuning of large-scale pre-trained language models. In this paper, we applied LoRA for text-guided single-image editing.

\section{Preliminaries}
\subsection{Problem Formulation}
Text-guided single-image editing aims to edit the given image of interest $I^{\text{input}}$ based on the target text prompts $P^{\text{tgt}}$ while keeping the remaining parts untouched. The general pipeline \cite{kawar2023imagic} to achieve this task is by leveraging the \textbf{text-to-image} diffusion models \cite{rombach2022high}, where the text condition comes from the CLIP's look-up table of word embeddings $P^{\text{tgt}}\rightarrow\mT^{\text{tgt}}\in\mathbb{R}^{77\times 512}$.  $77$ indicates the maximum number of words and $512$ is the embedding dimension. The core idea of diffusion models is to initialize with a randomly sampled noise image $\vz_T \sim \mathcal{N}(0,\mI)$ and gradually denoise it until a photorealistic image $\vz_0$. The image $\vz_t$ of time step $t$, where $t \in \{0,\ldots,T\}$ can be represented by:
\begin{equation}
\vz_t = \sqrt{\alpha_t}\vz_0 + \sqrt{1-\alpha_t}\rvepsilon_t,
\label{addnoise}
\end{equation}
where $0=\alpha_T<\alpha_{T-1}<\ldots<\alpha_1<\alpha_0=1$ are the hyper-parameters of the diffusion process, and $\rvepsilon_t \sim \mathcal{N}(0,\mI)$. The denoising steps are achieved by a network $f_{\theta}(\vz_t, t)$ conditioned on the image $\vz_t$ and word embedding $\mT$ so that noise $f_{\theta}(\vz_t,t,\mT)\approx \evepsilon_t$ can be faithfully sampled. To allow the diffusion model to fully adapt the given image from its original semantics $\mT^{\text{input}}$ to the target semantic $\mT^{\text{tgt}}$, the following two steps are required:

\vspace{3pt}
\noindent\textbf{A. Text embedding optimization}. \; The target text is first projected into the CLIP text embeddings $\mT^{\text{tgt}} \in \mathbb{R}^{77\times 512}$. To obtain the conditional text embeddings of the input image, we can take the target text embeddings $\mT^{\text{tgt}}$ as trainable parameters and \textbf{freeze} the parameters of the U-net in diffusion models. The encoder first projects the input image into the latent representation $\vz$, and optimizes the target text embedding $\mT^{\text{tgt}}$ using the denoising object:
\begin{equation}
\mathcal{L}(\vz,\mT^{\text{input}}, \theta) = \mathbb{E}_{t,\epsilon} \left[\lVert \epsilon - f_{\theta}(\vz_t,t,\mT^{\text{input}}) \rVert_2^2 \right],
\label{objective}
\end{equation}
where $t$ is uniformly sampled from $[1,1000]$, and $\vz_t$ is the noisy sample obtained with $\rvepsilon_t \sim \mathcal{N}(0,\mI)$  in Eq. \ref{addnoise}. The resulting $\mT^{\text{input}}$ is supposed to match the input image as closely as possible.

\vspace{3pt}
\noindent\textbf{B. U-net Fine-tuning}. \; Subsequently, the parameters $\theta$ of the U-net are\textbf{ fine-tuned} to further align the input image and the obtained text embedding $\mT^{\text{input}}$. With the same objective function in Eq. \ref{objective}, the U-net can learn to \textit{overfit} the input image of interest. Notably, the time step condition for each iteration is \textit{uniformly} sampled between 1 and 1000.
\begin{figure}[t]
    \centering
    \includegraphics[width=\linewidth]{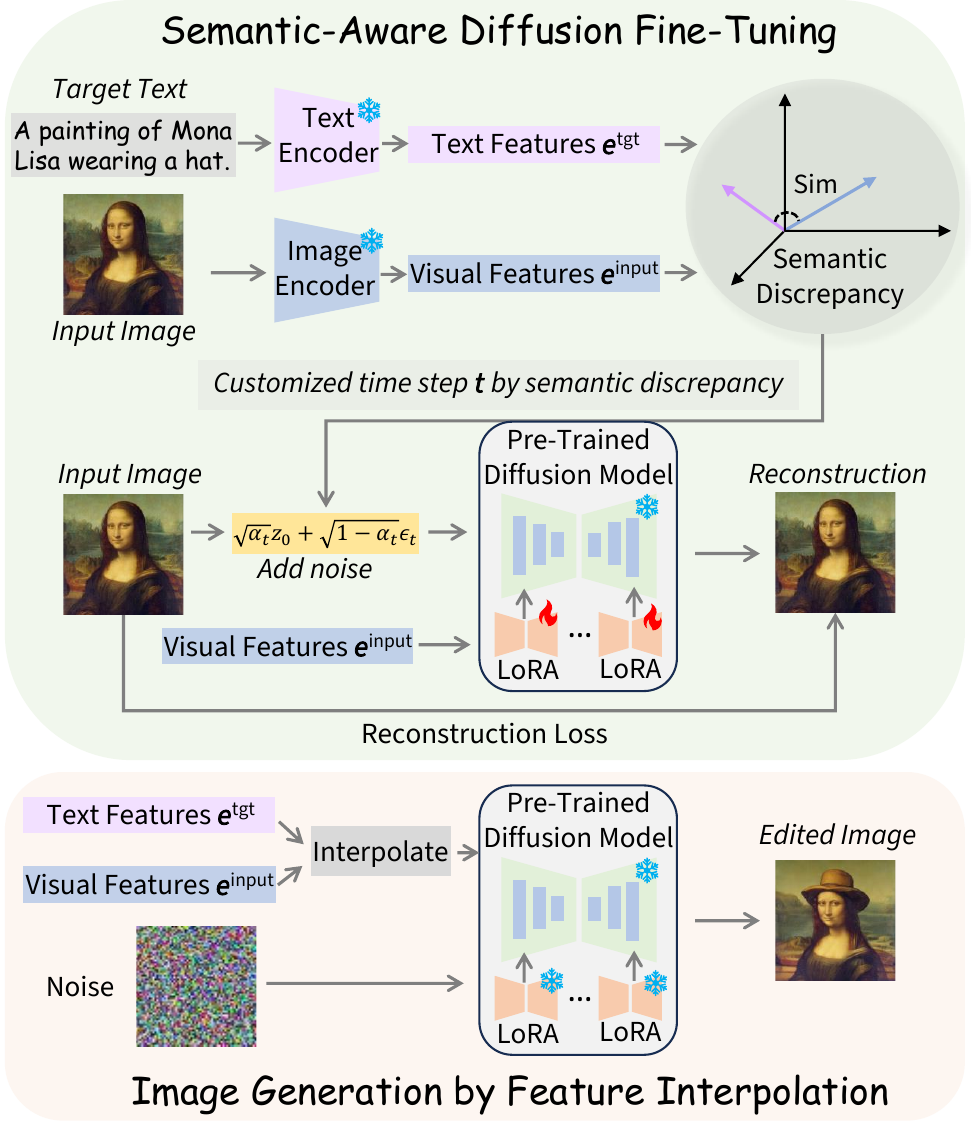}
    \vspace{-1.5em}
        \caption{Illustration of FastEdit. \textit{{\normalfont{Given an input image and a target text, we first project them into features using the CLIP model. Then, we calculate the semantic discrepancy between the two to determine the denoising time steps. Further, we fine-tune the low-rank matrixes added to diffusion model for a few iterations. Lastly, we can interpolate the CLIP's features to generate desired images with the fine-tuned model.}}}}
    \label{fig:arc}
    \vspace{-18pt}
\end{figure}

\subsection{Discussion}
To address the practical constraints of Imagic's time-consuming fine-tuning process, we need to tackle the significant time requirement. Currently, the process necessitates about 7 minutes per image due to its two-step fine-tuning approach. 
In Step A, text embeddings are optimized to align with the input image over 1,000 iterations, consuming roughly two minutes. Step B extends the process by an additional five minutes to fine-tune the entire U-net model over 1,500 iterations. To improve the efficiency and practicality of this procedure, we propose reevaluating the necessity of each step and exploring potential refinements in three key aspects:

\begin{itemize}[align=parleft,left=0em]
\item{ \textbf{[Text Optimization]} The two-minute duration required to optimize text embedding for each image seems \textit{unnecessary}, as we can project the input image and the target text into the same latent \textbf{feature} space. When applying the fine-tuned model to the same input image but a different target text, the text features and diffusion model can be reused without any adaptation.}
\item{\textbf{[Fine-tuning Steps]} The requirement of 1,500 iterations for fine-tuning the U-net, which typically consumes five minutes, poses a question about the possibility of reducing the number of training steps without compromising the quality of the output.}
\item{\textbf{ [Parameters]} The full fine-tuning of model parameters raises the issue of whether such an extensive approach is necessary. }
\end{itemize}





\begin{figure}[t]
    \centering
    \includegraphics[width=\linewidth]{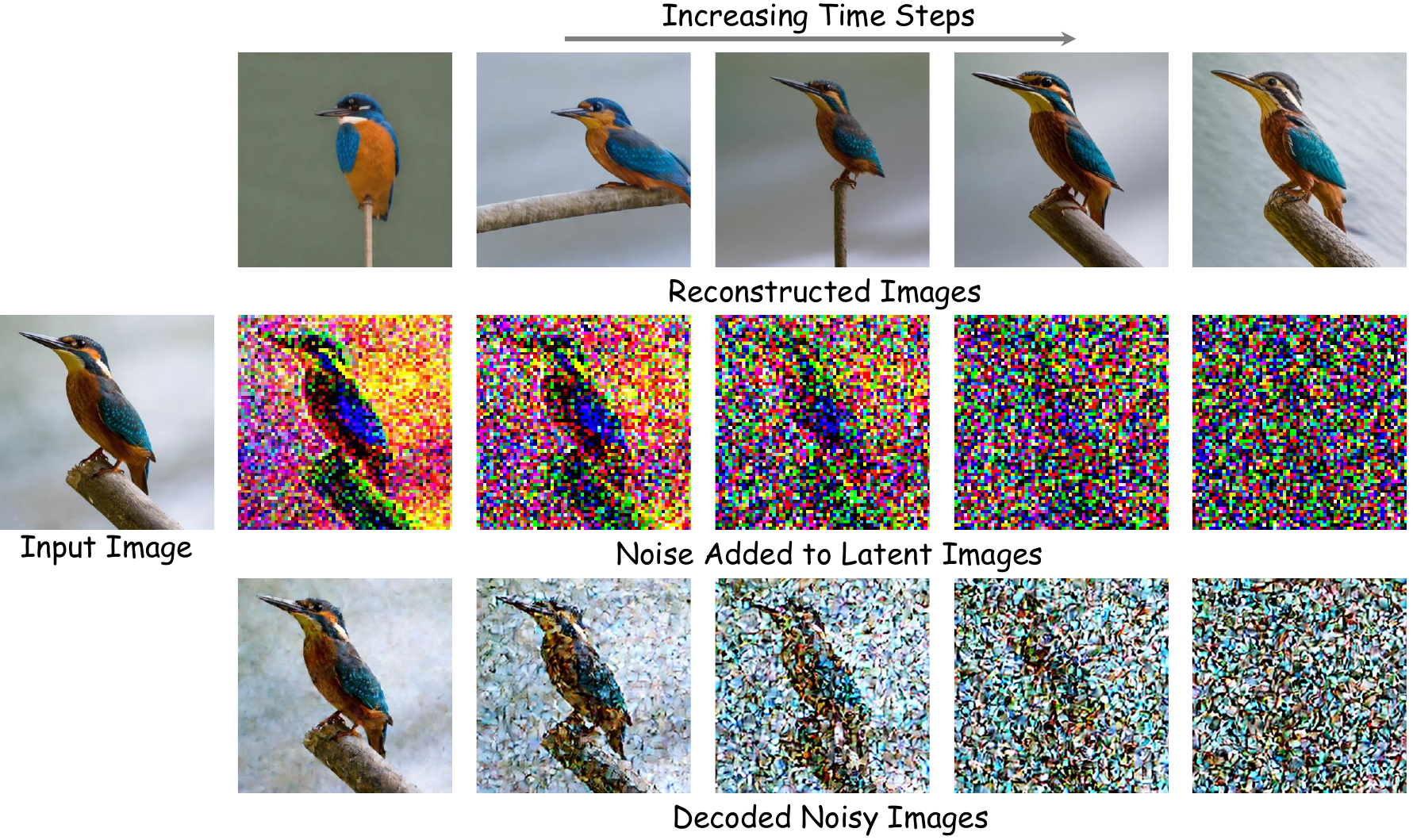}
    \vspace{-2.2em}
        \caption{Fine-tuning on a single denoising time step leads to texture and structure tradeoff. Low time step values tend to preserve the texture details of the object and result in input image structure distortion. In contrast, high time step values tend to preserve the image structure but lose textural details.}
    \label{fig:timesteps}
    \vspace{-10pt}
\end{figure}

\section{Proposed Approach: FastEdit}
With above motivations, in this section, we introduce FastEdit as shown in {\color{red}{Fig. \ref{fig:arc}}}. to accelerate the text-guided single-image editing process, which includes an image-to-image Stable Diffusion model, a semantic-aware diffusion fine-tuning strategy and parameter-efficient tuning with LoRA.

\subsection{Image-to-Image Diffusion Model}
The text-to-image variant generates images conditioning on the text embeddings $\mT \in \mathbb{R}^{77\times 512}$. We argue that this conditional information is not suitable for single-image editing task, as it does not fully exploit the CLIP's capacity of alignment between the textual and visual features. 
To solve this problem, drawing inspiration from CLIP model \cite{radford2021learning}, we employ an image-to-image variant to replace the text-to-image Stable Diffusion. In this way, we can simply use CLIP's image encoder and text encoder to extract the features of the input image and the target text:
\begin{equation}
\ve^{\text{input}} = \texttt{ImageEnc}(I^{\text{input}}), \:\:
\ve^{\text{tgt}} =  \texttt{TextEnc}(P^{\text{tgt}}),
\end{equation}
where $I^{\text{input}}$ and $P^{\text{tgt}}$ represent the input image and the target text. 
Thus, we can bypass the text embedding optimization step. 
Notably, with image-to-image variant, the visual features of the input image are not obtained by optimizing the target text, making it reusable for more editing target text prompts. Therefore, when more edits are required for the same image, the retraining of the text embedding step is also not necessary.

\subsection{Semantic-Aware Diffusion Fine-Tuning}
In the inversion process of diffusion models, the noise added to the input image is determined by a specific time step value. The time step is uniformly sampled between 1 and 1000, which is a large range. Thus, to fine-tune a diffusion model, it usually requires very large training iterations (1,500 for Imagic) to accommodate various time steps. This is the core challenge that hinders the acceleration of the editing process. 
We observe that the magnitude of the time step value significantly impacts the fine-tuning process of the model. As shown in {\color{red}{Fig. \ref{fig:timesteps}}}, the time step value controls the noise added to the image. We fine-tuned the diffusion model for multiple iterations to reconstruct the input image by conditioning on CLIP's input image features at a fixed time step $t$. Then, we can use the CLIP's input image features to reconstruct the input image. By increasing $t$, we observe that a small $t$ leads to object structure change, but the texture is well-preserved. In contrast, a large $t$ leads to texture loss but the structure and the composition is well-preserved.

Drawing inspiration from the above observation, we hypothesize that when the target text has a large semantic discrepancy to the input image, we are supposed to fine-tune the model with a large time step $t$ to keep the image structure from changing, and vice versa. 
Specifically, we first measure the semantic discrepancy between the input image and the target text:
\begin{equation}
\begin{gathered}
D(\ve^{\text{input}}, \ve^{\text{tgt}}) = 1 - \frac{\ve^{\text{input}} \cdot \ve^{\text{tgt}}}{\lVert\ve^{\text{input}} \rVert \: \lVert \ve^{\text{tgt}} \rVert}.
\end{gathered}
\end{equation}
With the obtained semantic discrepancy value, we customize a set of time steps $\{S\}$ with regard to $D(\ve^{\text{input}}, \ve^{\text{tgt}})$, where $\lVert S\rVert << 1000$.
Then, we fine-tune the diffusion model for 50 iterations by sampling $t$ from $\{S\}$. 
A few examples are shown in {\color{red}{Fig. \ref{fig:semantic_discrepancy}}}. When aiming to make a corgi dog standing, a low semantic discrepancy value is obtained. Then, We fine-tune the diffusion model for 50 iterations with low, medium, high and random time steps sets, respectively. After image editing, we observe that the subtle details in the input image can be changed if we use large time steps. 
Notably, the large time steps can even cause counterfactual results (a dog with five legs). In contrast, with a small set of time steps, we can achieve good editing results. Comprehensive experiments on various image-text pairs verified the effectiveness of this design. 

\subsection{Parameter-Efficient Fine-Tuning}
To further reduce the time cost, we resort to the parameter-efficient fine-tuning techniques, including prompt-tuning \cite{zhou2022learning}, adapter \cite{gao2024clip}, and low-rank adaptation (LoRA) \cite{hu2021lora}. In text inversion \cite{huang2023kv}, prompt-tuning is adopted to subject-driven image editing. When we adopt this technique for single-image editing, we observe that the target image is usually distorted from the input image. The similar issue is observed in \cite{ruiz2023dreambooth}, and referred as language shift problem. 
The adapter is also used for adding extra conditions to Stable Diffusion \cite{mou2024t2i}. However, when adapting it to single-image editing, we need to have a large number of parameters ($\sim$ 50\% of the U-net size) to achieve similar results as full fine-tuning. In comparison, low-rank adaptation (LoRA) \cite{hu2021lora} can reduce the parameters to 0.37\% of U-net's size. Specifically, we construct low-rank matrixes to all existing linear layers. Given a weight matrix $\mW \in \mathbb{R}^{d\times k}$ in U-net, which can be the feed-forward layers, or the self-attention layers, we introduce two low-rank matrixes $\mA \in \mathbb{R}^{d\times r}$ and $\mB \in \mathbb{R}^{r\times k}$, where $r<< min(d,k)$. The adaptation is then applied to the original weight matrix during the forward pass, formulated as $\bar{\mW} = \mW + AB$.
Here, $\bar{\mW}$ is the adapted weight matrix used in the computation. The original weight matrix $\mW$ remains frozen, and only the matrices $\mA$ and $\mB$ are updated.

\begin{figure}[t]
\vspace{-7pt}
\begin{algorithm}[H]
    \caption{The Pseudo-code of FastEdit}
    \renewcommand{\algorithmicrequire}{\textbf{Input:}}
    \renewcommand{\algorithmicensure}{\textbf{Output:}}
    \begin{algorithmic}[1]
        \REQUIRE A input image $I^{\text{input}}$, target text prompt $P^{\text{tgt}}$, CLIP model, image-to-image variant of Stable Diffusion model, hyperparameter $\{\alpha_0,
        \alpha_1,\ldots,\alpha_t\}$
        \ENSURE Target Image $I^{\text{tgt}}$
        \STATE $\{\ve^{\text{input}},\ve^{\text{tgt}}\}$ $\leftarrow$ Extract features from $I^{\text{input}}, P^{\text{tgt}}$ with CLIP
        \STATE $\text{D}(\ve^{\text{input}},\ve^{\text{tgt}})$ $\leftarrow$ Measure semantic discrepancy 
        \STATE $\{S\}$ $\leftarrow$ Customize a set of time steps with ${D}(\ve^{\text{input}},\ve^{\text{tgt}})$ 
        \STATE $\vz_{0}$ $\leftarrow$ Obtain the latent image from $I^{\text{input}}$
        \STATE Construct LoRA weights for the denoising U-net model \vspace{-10pt}
        \FOR {i = 1, 2, \ldots, 50 }
            \STATE Sample a time step condition $t$ from ${S}$
            \STATE Add noise to sample $\vz_t = \sqrt{\alpha_t}\vz_0 + \sqrt{1-\alpha_t}\rvepsilon_t$
            \STATE Fine-tune LoRA with $\mathbb{E}_{t,\epsilon} \left[\lVert \epsilon - f_{\theta}(\vz_t,t,\ve^{\text{input}}) \rVert_2^2 \right]$
        \ENDFOR
        \STATE $\bar{\ve}$ $\leftarrow$ Interpolate the input image features $\ve^{\text{input}}$ and the target text features $\ve^{\text{tgt}}$,  $\bar{\ve} = \eta\times \ve^{\text{tgt}} + (1-\eta)\times \ve^{\text{input}}.$
        \STATE $\bar{\vz}$ $\leftarrow$ Generate a latent image with $\bar{\vz}^{t-1} = f_{\theta}(\vz^t,t,\bar{\ve})$
        \STATE $I^{\text{tgt}}$ $\leftarrow$ Upsample $\bar{\vz}$ to high-resolution with SD decoder \vspace{-10pt}
        \RETURN $I^{\text{tgt}}$
    \end{algorithmic}
\end{algorithm}
\vspace{-30pt}
\end{figure}

\subsection{Image Generation by Feature Interpolation} 
To generate the edited image, 
we use the fine-tuned model to apply desired edits by advancing the direction of the  $\ve^{\text{tgt}}$. To do so, the linear interpolation between $\ve^{\text{input}}$ and $\ve^{\text{tgt}}$ is applied:
\begin{equation}
\bar{\ve} = \eta\times \ve^{\text{tgt}} + (1-\eta)\times \ve^{\text{input}}.
\label{interpolation}
\end{equation}
We then apply the generative diffusion process using the fine-tuned model conditioned on $\bar{\ve}$. This will result in the latent representation of the edited image $\bar{\vz}$, and the variational decoder can further produce the high-resolution image $I^{\text{tgt}}$.

\begin{table*}[tbp]
    \centering
    \caption{Comparison to Baseline Methods. We record the least time and iterations that successfully edit the majority of images.}
    \vspace{-2ex}
    \scalebox{0.9}{
    \begin{tabular}{r|r|c|c|c|c|c|c|c|c}
    \toprule
    \multirow{2}*{Methods} & \multirow{2}*{Venue} & CLIP \multirow{2}*{$\uparrow$} & \multirow{2}*{LPIPS $\downarrow$} & Human & \multirow{2}*{Iterations}  & \multirow{2}*{Time} & Trainable & Text  & Arbitrary    \\
    &  & Score $\;\;$ && Preference && & Parameters & Optimization   &  Text   \\
    \midrule
    LoRA \cite{hu2021lora} & ICML'21 & 0.232 & \textbf{0.349} & 9.7$\%$  &1000 & 101s   & 3.2M & \ding{51} & \ding{55} \\
    SVDiff \cite{han2023svdiff} & ICCV'23 & 0.247 & 0.453 & 23.8$\%$ & 1000 & 170s   & 0.2M & \ding{51} & \ding{55} \\
    Imagic \cite{kawar2023imagic} & CVPR'23 & 0.248 & 0.485 & 28.9$\%$ & 2500 & 7mins & 860M & \ding{51} & \ding{55}\\
    \midrule
    FastEdit&  & \textbf{0.269} & 0.524 & \textbf{37.6$\%$} & \textbf{50}   & \textbf{17s}   & \textbf{3.2M} & \ding{55} & \ding{51} \\
    \bottomrule
    \end{tabular}}
    \label{tablecompare}
    \vspace{-3pt}
\end{table*}


\vspace{-1ex}
\section{Experiments}
\subsection{Experimental Setup} 
\noindent \textbf{Datasets.} To better understand various approaches, we collect 54 image-text pairs from a wide range of domains, i.e., free-to-use high-resolution images from Pexels. We also conduct experiments on the TedBench (Textual Editing Benchmark) \cite{kawar2023imagic}, including 100 pairs of input images and target texts describing a desired complex non-rigid edit. 

\vspace{3pt}
\noindent \textbf{Implementation Details.} We implement FastEdit with the image variation of Stable Diffusion (based on Latent Diffusion Models \cite{rombach2022high}) in Python Diffusers \cite{von-platen-etal-2022-diffusers}. Note that the baseline method Imagic is applied to both pre-trained Imagen \cite{saharia2022photorealistic} and Stable Diffusion \cite{rombach2022high}. Due to the generative power difference, image editing performance is also better in Imagen. However, only Stable Diffusion provides open-source pre-trained models, we compare our results using Stable Diffusions.
The training iteration is set to 50. We adopt AdamW optimizer with learning rate 5e-4. We adopt DDIM  \cite{song2020denoising} as our denoising scheduler.  We have three sets of numbers for low, medium and high time steps: $\{200, 300, 400, 600\}$, $\{200, 400, 600, 800\}$ and $\{300, 500, 600, 800\}$ respectively. The editing process takes only 17 seconds on a single Tesla A6000. 

\vspace{3pt}
\noindent \textbf{Baselines.} We primarily compare our method to the state-of-the-art single-image editing method Imagic \cite{kawar2023imagic}. In line with the motivation of FastEdit, we also compare with other efficient fine-tuning methods. SVDiff \cite{han2023svdiff} is an efficient subject-driven image editing method, which only involves 0.2M trainable parameters. To enable single-image editing for SVDiff, we apply the same text embedding optimization from Imagic. As we adopt LoRA \cite{hu2021lora} to reduce trainable parameters, we further compare FastEdit with LoRA only, which does not involve image-to-image SD and semantic-aware diffusion fine-tuning. For all the methods, we search for the best hyperparameters for editing. Then, we sample 8 images using different random seeds and display the result with the best alignment to both the target text and the input image. 

\begin{figure}[t]
    \centering
    \vspace{5pt}
    \includegraphics[width=\linewidth]{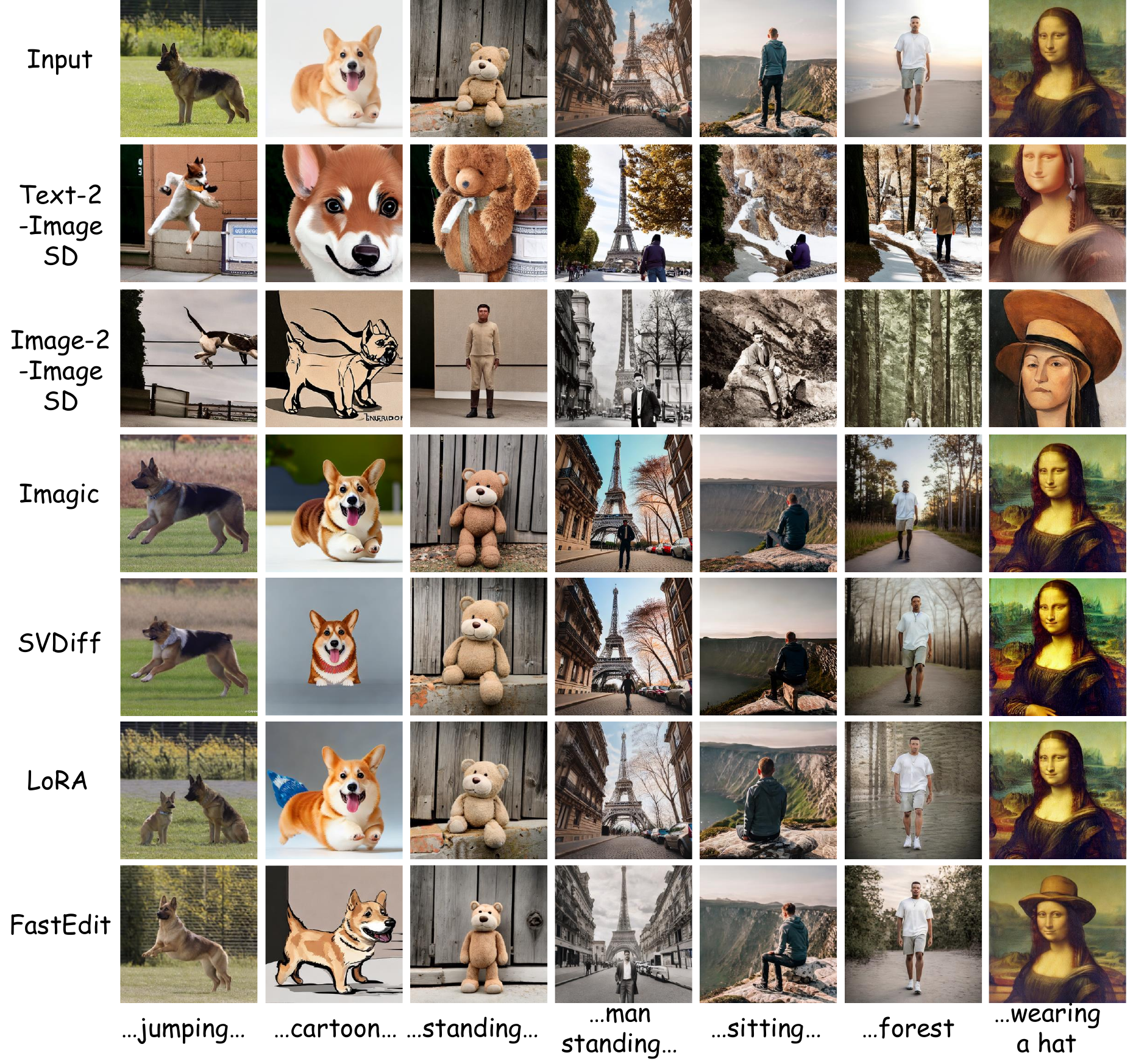}
    \vspace{-2.2em}
        \caption{Method comparison. We compare Imagic \cite{kawar2023imagic}, SVDiff \cite{han2023svdiff}, LoRA \cite{hu2021lora} with our method. FastEdit successfully applies the desired edit and preserves the original details.}
    \label{fig:comparison}
    \vspace{-10pt}
\end{figure}

\subsection{Qualitative Evaluation}
FastEdit can edit a single image with various target texts. As shown in {\color{red}{ Fig. \ref{fig:multi_texts}}}, we display editing results to various kinds of images, including animals, scene, human, and painting. From left to right, the target texts have larger semantic discrepancy from the original images. With our proposed semantic-aware diffusion fine-tuning, we ensure successful edits by applying different time steps on diffusion fine-tuning. Notably, in these images, we have applied edits including content addition, style transfer, background replacement, posture manipulation, \textit{etc}. Some impressive edits, such as transforming a corgi dog to a cat and making the Girl with a Pearl Earring gives a big grin, demonstrate FastEdit not only accelerate the editing time but also maintain high quality editing capacity.

\begin{figure*}[t]
    \centering
    
    \includegraphics[width=0.95\linewidth]{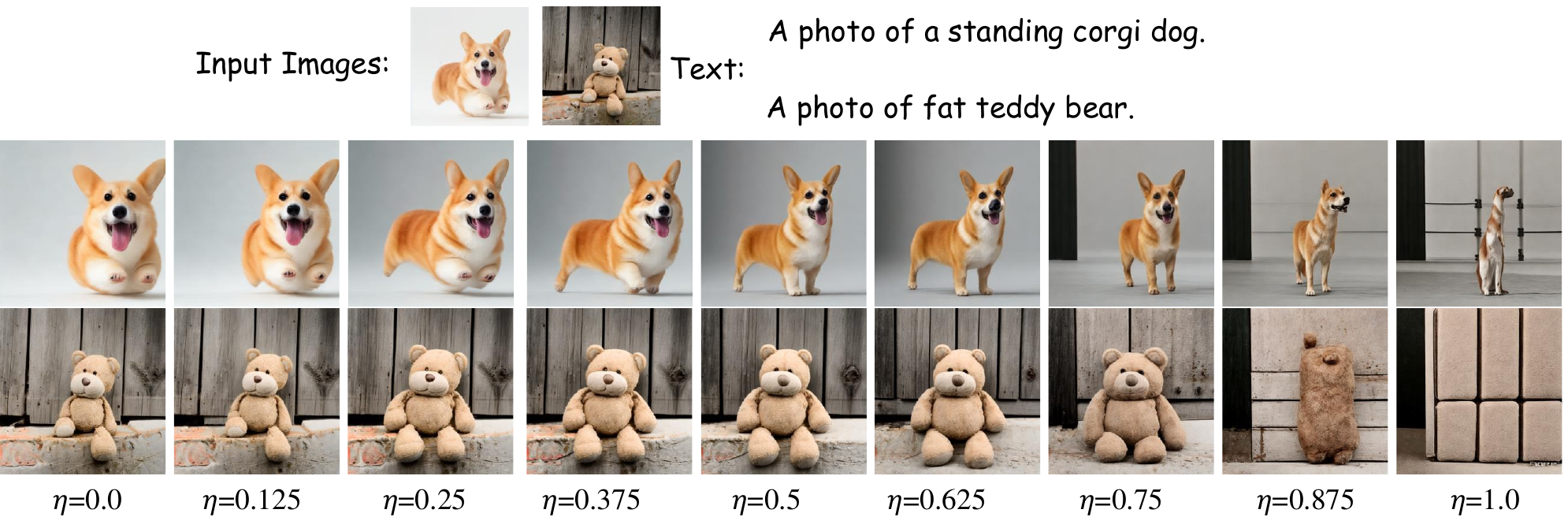}
    \vspace{-1.2em}
        \caption{Conditional feature interpolation. Increasing $\eta$ with the same seed, we see the resulted images are closer to the model's original understanding towards the target text.}
    \label{fig:interpolation}
    \vspace{-5pt}
\end{figure*}


\begin{figure}[t]
    \centering
    \includegraphics[width=\linewidth]{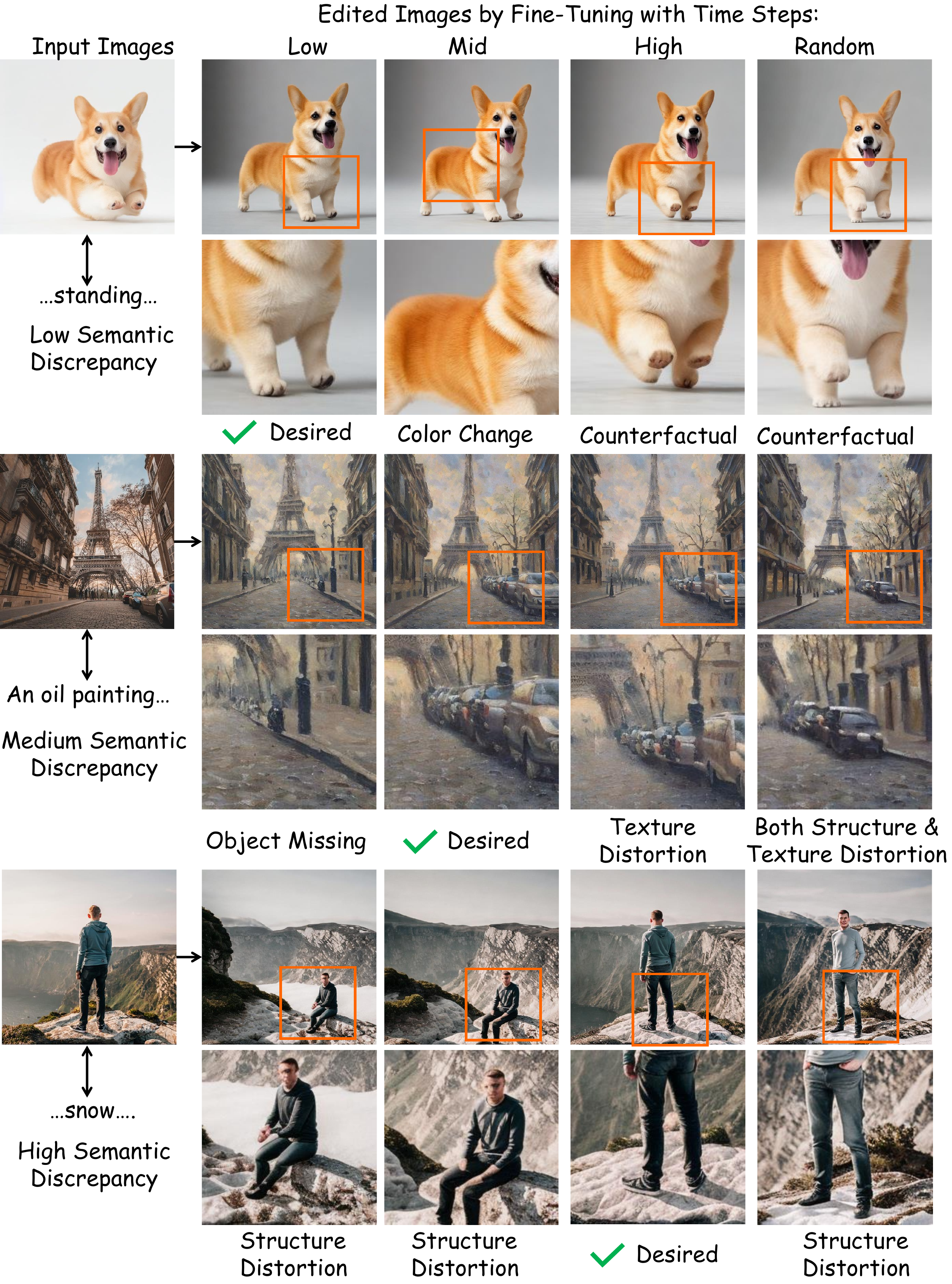}
    \vspace{-2.2em}
        \caption{Effects of semantic-aware diffusion fine-tuning. Different range of time step choices lead to different results on images with various semantic discrepancies. }
    \label{fig:semantic_discrepancy}
    \vspace{-15pt}
\end{figure}

\begin{figure}[t]
    \begin{minipage}{0.68\columnwidth}
        \includegraphics[width=\linewidth]{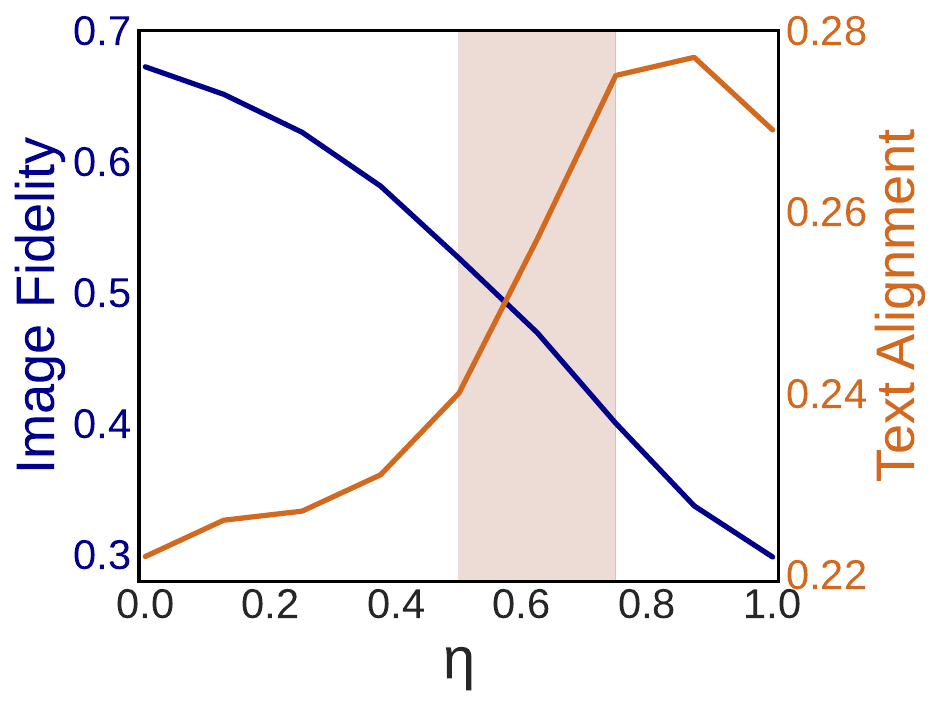} 
    \end{minipage}%
    \hfill%
    \begin{minipage}{0.3\columnwidth}
    \vspace{-20pt}
        \caption{Fidelity and alignment tradeoff. Edited images tend to match both the input image and text when $\eta$ is around 0.6.}
    \label{fig:sidecaption}
    \end{minipage}
    \vspace{-20pt}
\end{figure}



\subsection{Comparisons with SOTA Methods}
\noindent\textbf{Qualitative Comparison.}
As shown in {\color{red}{Fig. \ref{fig:comparison}}}, we randomly pick 7 image-text pairs to conduct text-guided single-image editing with the above 6 methods. Although the image-to-image SD performs the worst among them, FastEdit that equips with image-to-image SD still performs the best among all compared methods, which demonstrates the effectiveness of the proposed semantic-aware diffusion fine-tuning strategy. Thus, FastEdit constitutes the first demonstration of such fast text-based edits applied on a single real-world image, while preserving the original image details well.

\vspace{3pt}
\noindent\textbf{Quantitative Comparison.}
We also quantitatively compare FastEdit with other methods using evaluation matrixes, including CILP score, LPIPS, human preferences, and fine-tuning details, as shown in {\color{red} Table {\ref{tablecompare}}}. CLIP score is calculated between the target text features and the edited images, which is also called text alignment. We achieve the best CLIP score among the compared methods. LPIPS \cite{zhang2018unreasonable} is the learned perceptual image patch similarity, which measures the distortion of edited images from the input image. However, we notice that the many failed cases in LoRA are usually reconstructed images of the input. Thus, LoRA has the least CLIP score, but also the least distortion from the input image. Compared with the reasonable LPIPS score in Imagic, we have similar editing success rate, and also similar LPIPS, which demonstrates our method can achieve balance between CLIP score (text alignment) and LPIPS (Image Fidelity).

\vspace{3pt}
\noindent\textbf{User Study.}
We have developed a comprehensive survey encompassing a broad range of image-text pairs to conduct a user study aimed at evaluating human preferences for text-guided image editing outputs. This study involves 120 participants who are presented with 54 pairs of input images and corresponding target texts. Each participant is tasked with selecting their preferred anonymous editing results from four different methodologies: Imagic, SVDiff, LoRA, and FastEdit, presented in multiple-choice questions. The collected preference scores are detailed in {\color{red}Table {\ref{tablecompare}}}, where we observe a pronounced preference for FastEdit among the participants. 



\vspace{-5pt}
\subsection{Ablation Study}
There are mainly three components in FastEdit to accelerate the editing procedure yet improve the editing performance, including the image-to-image variant, semantic-aware diffusion fine-tuning, and LoRA. To first prove that our editing performance is not attributed from a better diffusion model, we compare the generated samples from text-to-image and image-to-image variants, respectively. As shown in {\color{red}{Fig. \ref{fig:comparison}}}, given the target texts, we extract the text embeddings and text features, and then use them to generate images accordingly. It is clear that the images from text-to-image SD can achieve significantly better performance. Thus, the adoption of image-to-image variant does not contribute to our editing performance.
Second, to showcase the effects of semantic-aware diffusion fine-tuning, we compare the edited images by fine-tuning with different time steps. As shown in {\color{red}{Fig. \ref{fig:semantic_discrepancy}}}, For example, the CLIP features of the target text "A photo of a standing corgi dog." have low semantic discrepancy from the input image. Thus, we suppose to fine-tune the model with low time steps, 
Third, we compare FastEdit with LoRA only in {\color{red}{Fig. \ref{fig:comparison}}} and {\color{red} Table {\ref{tablecompare}}}. LoRA only is the variant that we add low-rank matrixes as trainable parameters to U-net. All other details are based on Imagic. It can be seen that with LoRA only, the edited images cannot faithfully reflect the semantics in the target text, but the number of trainable parameters is reduced to 0.37\% of U-net's size.

\subsection{Feature Interpolation Thresholding}
The hyper-parameter $\eta$ controls the interpolation between the input image and the target text. As shown in {\color{red}{Fig. \ref{fig:interpolation}}}, employing a small $\eta$ results in minimal deviations from the input image. Conversely, When utilizing a significantly larger $\eta$ leads to more pronounced changes, at times resulting in distortions that diverge from realistic modifications. This behavior likely stems from a mismatch between the image features, which the image-to-image variant of the model is trained on. Thus, the target text features can introduce discrepancies. 
As shown in {\color{red}{Fig. \ref{fig:sidecaption}}}, we present the image fidelity (1-LPIPS) and the text alignment (CLIP score) trend while increasing $\eta$, an $\eta$ value of approximately 0.6 strikes a balance, enabling meaningful yet coherent transformations aligned with the textual guidance without compromising the image's realism. 

\section{Conclusion}
In this paper, we propose a fast text-guided single-image editing approach via semantic-aware diffusion fine-tuning, which only requires 17 seconds per image. To speed up the editing process, we replace the conventional text-to-image variant with an image-to-image variant that circumvents the text embedding optimization step, which typically requires 2 minutes per image. Then, our semantic-aware diffusion fine-tuning can customize time step choices to trade off texture and structure in the edited images and reduce the original 1,500 iterations to 50 only. Further, we apply low-rank adaptation to reduce the number of trainable parameters to 0.37\% of U-net's size. Extensive experiments demonstrate the efficiency and effectiveness of FastEdit.

{\small
\bibliographystyle{ieee_fullname}
\bibliography{egbib}
}

\end{document}

%% file: math.tex
\usepackage{amsmath,amsfonts,bm}

\def\eqref#1{equation~\ref{#1}}

\def\1{\bm{1}}

\def\rvepsilon{{\mathbf{\epsilon}}}

\def\ve{{\bm{e}}}

\def\vz{{\bm{z}}}

\def\evepsilon{{\epsilon}}

\def\mA{{\bm{A}}}
\def\mB{{\bm{B}}}

\def\mI{{\bm{I}}}

\def\mT{{\bm{T}}}

\def\mW{{\bm{W}}}

\DeclareMathAlphabet{\mathsfit}{\encodingdefault}{\sfdefault}{m}{sl}
\SetMathAlphabet{\mathsfit}{bold}{\encodingdefault}{\sfdefault}{bx}{n}

